\documentclass[times,twocolumn,final]{elsarticle}

\usepackage{cag}
\usepackage{framed,multirow}

\usepackage{amssymb}
\usepackage{latexsym}

\usepackage{url}
\usepackage{xcolor}
\definecolor{newcolor}{rgb}{.8,.349,.1}

\usepackage{hyperref}

\journal{Computers \& Graphics}

\begin{document}

\verso{Preprint Submitted for review}

\begin{frontmatter}

\title{PIG-Net: Inception based Deep Learning Architecture for 3D Point Cloud Segmentation}%

\author[1]{Sindhu \snm{Hegde}
}
\ead{sindhu.hegde@research.iiit.ac.in}
    
\author[2]{Shankar \snm{Gangisetty}\corref{cor1}
}
\cortext[cor1]{Corresponding author 
 }
\emailauthor{shankar@kletech.ac.in}{Shankar Gangisetty}

\fntext[fn1]{\copyright {2021}. This manuscript version is made available under the CC-BY-NC-ND 4.0 license \url{http://creativecommons.org/licenses/by-nc-nd/4.0/}\\
DOI: \url{https://doi.org/10.1016/j.cag.2021.01.004}
}  

\address[1]{IIIT Hyderabad, India}
\address[2]{KLE Technological University, Hubballi, India}

\received{January 12, 2021}

\typeout{get arXiv to do 4 passes: Label(s) may have changed. Rerun}

\begin{abstract}
Point clouds, being the simple and compact representation of surface geometry of 3D objects, have gained increasing popularity with the evolution of deep learning networks for classification and segmentation tasks. Unlike human, teaching the machine to analyze the segments of an object is a challenging task and quite essential in various machine vision applications. In this paper, we address the problem of segmentation and labelling of the 3D point clouds by proposing a inception based deep  network architecture called PIG-Net, that effectively characterizes the local and global geometric details of the point clouds. In PIG-Net, the local features are extracted from the transformed input points using the proposed inception layers and then aligned by feature transform. These local features are aggregated using the global average pooling layer to obtain the global features. Finally, feed the concatenated local and global features to the convolution layers for segmenting the 3D point clouds. We perform an exhaustive experimental analysis of the PIG-Net architecture on two state-of-the-art datasets, namely, ShapeNet~\cite{Yi:2016:ShapeNet} and PartNet~\cite{Mo:2019:PartNet}. We evaluate the effectiveness of our network by performing ablation study. 
\end{abstract}


\end{frontmatter}


\section{Introduction}
\label{introduction}
Human vision is the most interesting and brilliant characters of human that  segments, categorizes, recognizes and organizes the objects based on the knowledge of their parts. In our world, human takes a critical decision based on the detection and reasoning of the parts in an object. For instance, if a human wishes to sit on a chair, the human vision detects the parts of the chair such as seat, arm, chair back, legs and reasons appropriately whether to sit or not. One of the reasoning is whether the legs of the chair are thin, broken, small or in an uncertain state to make a critical decision. Unlike human vision, supervising the machine vision to analyze the parts of an object i.e., segmentation and labeling is a challenging task. Machine vision segmentation is essential in many applications like robot navigation and grasping~\cite{Max:2018:IJRR,Rusu:2013,Rao:2010:IRS}, generating 3D shapes~\cite{Wu:2018,Li:2017,Zou:2017:3DPRNNGS}, medical surgery~\cite{Wang:2018:InteractiveMI, Xuan:2018:Brain}, 3D modeling and animation~\cite{Xu:2012:FDS,Fu:2017:TVCG,Ichim:2017:PPF}, and scene understanding~\cite{Sakaridis:2018:ModelAW,Zheng:2015:IJCV}.

Initially most of the works in literature adopted hand-crafted feature-based learning approaches for segmentation and labeling~\cite{WangCG:2018:ShapeSeg,Schneider:2016:ESS,ZhangCG:2015:SplatSeg,Xie:2014:ELM,NIPS:2011,Xu:2003,Kalogerakis:2010} on small datasets, this  misses the fine-grained details for reasoning over the different parts in an object. Recently, with large and growing online repositories of data, deep learning has become a general tool for many of the computer vision tasks ranging from classification to segmentation to scene understanding, especially convolutional neural networks (CNN) in 2D images~\cite{Farabet:2013:LHF,Simonyan:2015:ICLR,He:2016:CVPR,Noh:2015:LDN,Long:2015:CVPR}. Since robotic vision is rapidly growing to automate every job (eg., industry to healthcare), currently most of the researchers aim at adaptation of deep CNN to 3D objects. Thus, the availability of large 3D object datasets with part annotations~\cite{Yi:2016:ShapeNet,Mo:2019:PartNet} and deep learning techniques, enables us to research on 3D object segmentation and labeling for object understanding.
Among the popular categorizes of point cloud based segmentation, namely, semantic segmentation at scene level~\cite{Hu:SemSeg:2020,Feng:SemSeg:2020,Ku:shrecsemseg:2020,Wang:InstSemaSeg:2019,Landrieu:SemSeg:2018,HuCG:SemSeg:2020,BoulchCG:SemSeg:2018,Zhao:SemSeg:2020,Akadas:2020:ACCV}, instance segmentation at object level~\cite{Jiang:InstanceSeg:2020,Han:OccuSeg:2020,Zhang:ICRA:2020,Yang:InstanceSeg:2019,Zhao:InstSeg:2020,Shih:InstanceSeg:2020}, and part segmentation at part level~\cite{Marios:PartSeg:2020,Thomas:2019:KPConv,Su:2018:Spaltnet,Xie:PartSeg:2018,Xu:2018:SpiderCNN,Lin:2020:3D-GCN}, in this paper we focus on 3D object part segmentation.

The 3D object segmentation and labeling task is non-trivial due to the representations of 3D data. The 3D objects captured using sensory devices are raw point clouds that are irregular and the online standard dataset repositories are mesh models that are moreover irregular. However, the CNN architectures require regular data formats in order to detect local features by weight-sharing and kernel machine optimization for computational efficiency. Most of the researchers transform point clouds and mesh models to 3D voxel grids or multi-view images before giving the input to deep CNN architecture. Though the volumetric grid is regular, it is computationally expensive and uses distance field as a means for data representation that itself is challenging to set depending on the nature of the 3D objects. In this paper, we focus on point clouds that are simple, homogeneous, expressive, compact representation of surface geometry, and avoids combinatorial irregularities and complexities of meshes for learning. We name our proposed point cloud network architecture as Point Inception Global average pooling network, \textit{PIG-Net}.

Our PIG-Net is a deep learning architecture that  directly takes point clouds as input and provides resulting output as per point segment labels for each point of the point clouds. The PIG-Net architecture is a combination of the input transform and the feature transform layers, inception module, global average pooling (GAP) layer and convolution layers. Initially, we adopt the input transformation of PointNet~\cite{Qi:2017:PointNet} to align all the input points. We then extract the local features using a series of proposed inception layers to  capture the fine-grained details, followed by feature transformation. To obtain the global features, these features are aggregated using the GAP layer that is robust to spatial translations of the point clouds and additionally avoids overfitting. The obtained local and global features are then concatenated and finally given to the convolution layers that segments the different parts of the 3D objects.    

The major contributions of our work are as follows:
\begin{itemize}
\item We propose a point inception based deep neural network called \textit{PIG-Net} for segmentation and labeling of the 3D point clouds.
\item To extract the local features, we propose inception layers based on  inception module as the intermediate layer along with input and feature transformation to extract the discriminative features in order to  improve the performance. In addition, we propose to use GAP over other pooling methods to avoid overfitting.
\item We provide an exhaustive evaluation and comparative analysis of  PIG-Net with the existing methods on two state-of-the-art datasets, namely, ShapeNet-part~\cite{Yi:2016:ShapeNet} and PartNet~\cite{Mo:2019:PartNet}. Additionally, the ablation study demonstrate the effectiveness of proposed PIG-Net.
\end{itemize}

In Section~\ref{related_works}, we review some of the existing segmentation works. In Section~\ref{methodology}, we discuss the proposed PIG-Net architecture. In Section~\ref{results}, we present the experimental results and analysis of the proposed network on state-of-the-art datasets. In Section~\ref{conclusions}, we provide concluding remarks.



\section{Related work}
\label{related_works}
We categorize the 3D object segmentation into two classes, namely, handcrafted feature learning and deep feature learning for point clouds.

\subsection{Handcrafted feature learning}
The traditional works rely on the handcrafted feature extraction from the 3D objects for specific data processing tasks. The point feature methods capture the shape of objects, the inherent geometric properties of points, and are invariant to certain transformations. The widely used feature descriptors can be classified into signature, histogram and transformation based descriptors~\cite{Guo:FeatureSurvey}. The signature based descriptors like normal aligned radial feature (NARF)~\cite{Steder:NARF} and normal based signature (NBS)~\cite{Li:NBS} are popular on range images. The histogram based descriptors like spin images~\cite{Johnson:SpinImages}, shape context~\cite{Frome:ShapeContext}, rotational projection statistics (RoPS)~\cite{Guo:RoPS}, 3D scale invariant feature transform (SIFT)~\cite{Scovanner:SIFT}, and fast point feature histogram (FPFH)~\cite{Rusu:FPFH} are popularly used in combinations of multi-view images and point clouds. The transformation based descriptors like 3D speeded up robust feature (SURF)~\cite{Lopez:3DSP} and spectral descriptors such as Laplace-Beltrami operator (LBO)~\cite{Belkin:LBO}, heat kernel signature (HKS)~\cite{Sun:HKS}, scale invariant HKS (SIHKS)~\cite{Bronstein:SIHKS}, wave kernel signature (WKS)~\cite{Aubry:WKS}, improved WKS (IWKS)~\cite{Frederico:Encoding} and metric tensor and Christoffel symbols~\cite{Setty:2018:EIP} have been popularly used on mesh and point clouds. After feature extraction, machine learning techniques are applied on these features either directly or by employing certain methods like patch clustering~\cite{Hu:2018} or feature encoding~\cite{HegdeG:2019}. However, the selection of optimal handcrafted features is non-trivial and highly data specific task as these approaches are often based on certain prior assumptions w.r.t. 3D objects.

\subsection{Deep feature learning}
Recent advances in deep learning tools have eliminated the need for hand-crafting the features. Many works have evolved on point clouds after a pioneering work of PointNet~\cite{Qi:2017:PointNet} making it possible for directly using point clouds.

In PointNet~\cite{Qi:2017:PointNet}, authors use $(x,y,z)$ coordinates of points as input features with multi layer perceptrons (MLPs) for classification and segmentation. These features are aggregated using the max pooling and forms global features. The network learns to summarize the input point cloud by a sparse set of key points, that roughly corresponds to the skeleton of the objects. The limitation is, this architecture does not capture the local structures induced by the metric. Exploiting such local structures has proven to be important for the success of convolutional architectures. Thus, as an extension to this network, authors in PointNet++~\cite{Qi:2017:PointNet++} proposed a hierarchical features architecture, that learns  effectively by adopting max pooling. But both the networks rely on max pooling for aggregation of local and global features. 

Unlike PointNet++~\cite{Qi:2017:PointNet++}, authors in SPLATNet~\cite{Su:2018:Spaltnet}, SFCNN~\cite{Rao:2019:SFCNN}, RIConv~\cite{Zhang:2019:RIConv} capture local properties differently by mapping original points into a  space lattice and processing using bilateral convolutional layers to improve stability. In SFCNN~\cite{Rao:2019:SFCNN}, the model also exploits  spherical lattice structure for generalization. However, in RIConv~\cite{Zhang:2019:RIConv}, authors do not use spherical lattice structure for rotation in-variance. Instead, they defined convolution with rotation invariant features.
These works explore both convolution on local point features and rotation invariance. But, the local geometric features are not complete because original point cloud coordinates are not retained in low-level geometric feature extraction, trading for rotation invariance.
In KCNet~\cite{Shen:2018:KCNet}, authors proposed a kernel correlation layer to improve PointNet by efficiently exploring local 3D geometric structures and local high-dimensional feature structures. The KCNet is based on graph representation. In KdNet~\cite{Klokov:2017:KdNet, Gadelha:2018:MRTNet}, authors extract the hierarchical features from the input point cloud using kd-tree. For segmentation, in KdNet~\cite{Klokov:2017:KdNet} the encoder-decoder architecture is mimicked with skip connections. But  due to the lack of overlapped receptive fields and the insufficient information propagation across the kd-tree structure, the performance is poor.

\begin{figure*}
\centering
\includegraphics[width=1\linewidth]{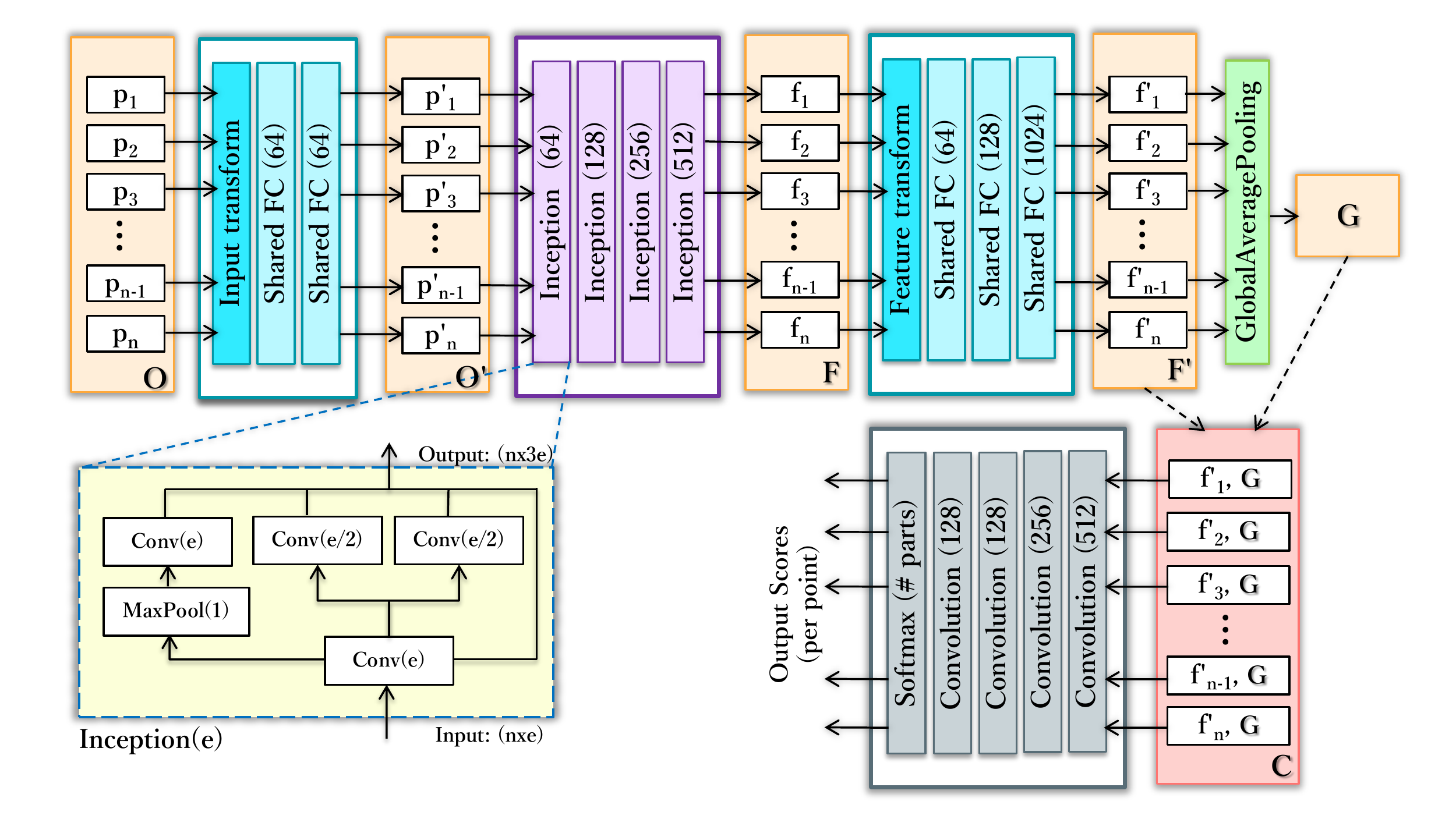}
\caption{Proposed inception based deep learning architecture of our PIG-Net.}
\label{architecture}
\end{figure*}

To overcome the issues of PointNet++ and Kd-Net, authors in SONet~\cite{Li:2018:SONet} propose a permutation invariant architecture. A Self-Organizing Map (SOM) is constructed which models the spatial distribution of the input point clouds. The hierarchical features are extracted from both the individual points and the SOM nodes. For segmentation of 3D objects, the global feature vector is directly expanded and concatenated with the normalized points. But the SONet suffers similar fails as that of PointNet. Another work to overcome the limitation of PointNet++ is PointGrid~\cite{Le:2018:PointGrid}. In this architecture, the input point cloud is embedded into a 3D grid. The 3D convolutional network extracts the features and learns higher order local approximation functions that captures the local geometry of the 3D shapes. For segmentation, the network employs the feature extraction and decodes the extracted features using deconvolution.
The limitation is random sampling that fails for large cluster value and is computationally expensive.

Unlike aforementioned methods there are other category of methods 
which learn the local properties from subset of 3D points~\cite{Wang:2019:DGCNN,Thomas:2019:KPConv,Lin:2020:3D-GCN} rather than the entire 3D object.
In DGCNN~\cite{Wang:2019:DGCNN}, authors propose EdgeConv to capture the local features by generating edge features which provides the relationship between points and the nearest neighbors. The EdgeConv is designed to be permutation-invariant, but the DGCNN is not effective during scale invariance and coordinate shifts are performed. The coordinate shifts occurs by extracting features from global coordinate. In KPConv~\cite{Thomas:2019:KPConv}, authors select neighboring points for each point by describing the local structures of point clouds with a radius, but KPConv cannot deal with scale variations. To overcome these drawbacks, authors in 3D Graph Convolution Networks (3D-GCN)~\cite{Lin:2020:3D-GCN}, present a learnable 3D graph kernels and graph max-pooling operation, to obtain geometric features across scales while exhibiting scale and shift-invariance.

Other segmentation methods like PointCNN~\cite{Li:2018:PointCNN}, SPH3D-GCN~\cite{Lei:2020:SPH3D-GCN} and PointConv~\cite{Wu:2019:PointConv} have complex convolution architectures to preserve local features. In this paper, we use inception module for preserving low-level features and GAP to  build a PIG-Net architecture for 3D point cloud part segmentation, yielding better performance.


\section{3D Point Cloud Segmentation}
\label{methodology}




We propose a new inception based deep network architecture called Point Inception Global average pooling network (\textit{PIG-Net}) shown in Figure~\ref{architecture}, in order to segment the different parts of the 3D point cloud. The major components of PIG-Net include, input transform and feature transform layers, inception module, global average pooling layer and convolution layers.
 Given a point cloud $O = \{p_i | p_i \in R^3, i = 1, 2, ..., n\}$, the input points $p_i$ are transformed 
by applying a symmetric function for permutation invariance and aggregate to obtain the  transformed points $O^{\prime} = \{p^{\prime}_1, p^{\prime}_2, ..., p^{\prime}_n\}$. The input transform network of PointNet~\cite{Qi:2017:PointNet}  is used to predict the affine transformation matrix (T-Net),  which aligns all the input points to a canonical space and makes the network to be invariant to certain geometric transformations, such as rigid transformation.

\subsection{Convolutions on Point Clouds}
The convolutions that are used in all the components of the PIG-Net architecture is defined as,
\begin{equation}
    \label{conv_def}
    Conv(q) = \sum_{j=1}^{d} w_jh_j
\end{equation}
where $Conv(q)$ is the convolution over each point $q$, $w_j$ is the $j^{th}$ component of the convolutional kernel weights $W \in R^d$ represented as a series of $1 \times d$ kernel weight vectors where $d$ is the number of dimensions and $h_j$ is the $j^{th}$ component of feature vectors $H \in R^d$.


\subsection{Inception module} We propose inception layers based on inception module that takes the transformed points $O^{\prime}$ as the input and outputs the local features $F = \{f_1, f_2, ..., f_n\}$ to capture the fine grained details of the points. In deep neural networks, one of the most common way to improve the performance is to increase either the number of layers or the number of neurons in each layer. But this results into overfitting due to the large number of parameters and furthermore increases the computational complexity due to the uniform increase in the filters. In order to improve the performance and to overcome these drawbacks, we propose an inception module for point clouds. The inception module~\cite{Szegedy:2015:Inception} had emerged as a breakthrough solution for classification and detection tasks in the ImageNet ILSVRC 2014 challenge.
Though the PointNet considers permutation invariance of unordered point sets in order to improve the performance, however the performance can be further enhanced by considering the optimal local sparse structure in the network. Thus, we use inception module as the intermediate layers in the PIG-Net as depicted in Figure~\ref{architecture} to enhance the network performance.

The proposed inception module consists of convolution layers and max pooling layer as shown in Figure~\ref{architecture}. Each convolution layer includes batch normalization to reduce the co-variance shift and rectified linear unit (ReLU) activation function adding non-linearity to the network and minimizes the vanishing gradient problem. The convolution layer identifies the basic patterns and concatenating several of these layers, creates hierarchical filters that represent the different parts of the point cloud. As illustrated in Figure~\ref{architecture}, the first convolutional layer convolves the input using $e$ filters. The output of this layer is fed to two convolution layers, each with $e/2$ filters and a max-pooling layer. The max-pooled features are given to another convolution layer with $e$ filters. Finally, the output from all the four convolution layers are concatenated to extract the local features. In our network, we use a series of four inception layers, starting with 64 filters which are doubled in the successive layers (i.e., 128, 256, 512). The extracted inception features precisely capture the shape of the point clouds by recognizing the fine-grained details.

These features are then aligned using the symmetric function to obtain the transformed features $F^{\prime} = \{f^{\prime}_1, f^{\prime}_2, ..., f^{\prime}_n\}$. A feature transformation matrix is predicted which aligns the features $F$ using an alignment network. Since the dimension of the transformation matrix is high, a regularization term is added to stabilize the optimization. The alignment network helps to make the learnt inception features be invariant to the geometric transformations and further improves the performance.

\subsection{Global average pooling  (GAP)} 
\label{lab:GAP}
We propose to use GAP~\cite{Min:2014:NiNGAP} layer to extract the global features $G$, characterizing the entire object. Conventionally, max-pooling that considers the maximum value to obtain the global features is adopted in majority of the networks including PointNet~\cite{Qi:2017:PointNet} and PointGrid~\cite{Le:2018:PointGrid}. Similar to max-pooling, GAP layer can be used to reduce the spatial dimensions and obtain a single global feature. However, since there are no parameters to be optimized, the GAP avoids overfitting of data and is additionally robust to spatial translations of the input point clouds. This is a potential advantage when compared to max-pooling aggregation. Thus, in our network, we propose to use GAP to obtain the global features. In CNNs, especially in images~\cite{Min:2014:NiNGAP}, GAP is used to replace the fully connected layers in the network. We use GAP in PIG-Net, not only to obtain the global features, but to provide a more native way to the convolution structure enforcing the correspondences between the features and parts of point cloud.

\begin{figure*}
\centering
\includegraphics[width=1\linewidth]{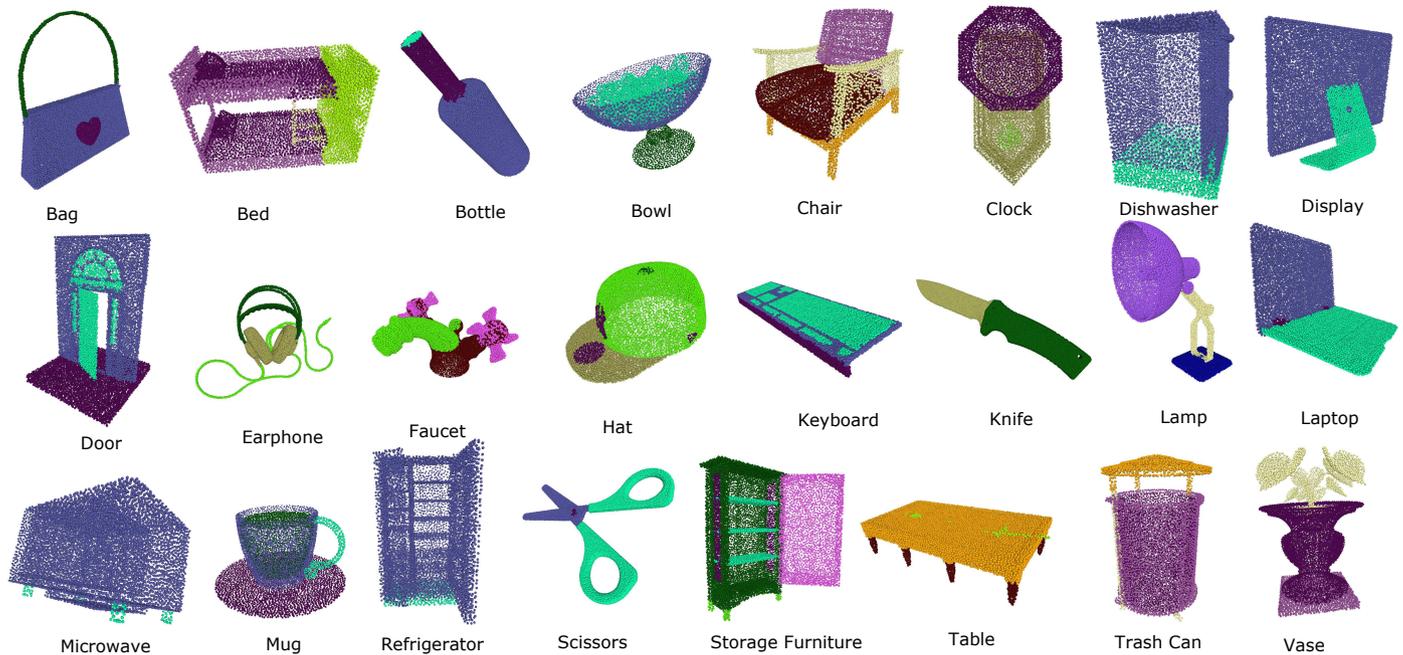}
\caption{Sample 3D point clouds from the PartNet dataset (coarse level)~\cite{Mo:2019:PartNet}.}
\label{samples}
\end{figure*}

\begin{table*}[ht]
\footnotesize
\setlength{\tabcolsep}{4.8pt}
\centering
\caption{Dataset statistics of ShapeNet-part dataset.}
\begin{tabular}{l*{17}{l}l}
\hline

\multicolumn{1}{l|}{} & \multicolumn{1}{c|}{Total} & aero & bag & cap & car & chair & ear & guitar & knife & lamp & laptop & motor & mug & pistol & rocket & skate & table\\
\multicolumn{1}{l|}{} & \multicolumn{1}{c|}{} &  &  &  &  &  & phone &  &  &  &  & bike &  &  &  & board & \\

\hline

\multicolumn{1}{l|}{\# parts} & \multicolumn{1}{l|}{50} & 4 & 2 & 2 & 4 & 4 & 3 & 3 & 2 & 4 & 2 & 6 & 2 & 3 & 3 & 3 & 3\\

\hline

\multicolumn{1}{l|}{\# train} & \multicolumn{1}{l|}{12137} & 1958 & 54 & 39 & 659 & 2658 & 49 & 550 & 277 & 1118 & 324 & 125 & 130 & 209 & 46 & 106 & 3835\\

\multicolumn{1}{l|}{\# validation} & \multicolumn{1}{l|}{1870} & 391 & 8 & 5 & 81 & 396 & 6 & 78 & 35 & 143 & 44 & 26 & 16 & 30 & 8 & 15 & 588\\

\multicolumn{1}{l|}{\# test} & \multicolumn{1}{l|}{2874} & 341 & 14 & 11 & 158 & 704 & 14 & 159 & 80 & 286 & 83 & 51 & 38 & 44 & 12 & 31 & 848\\

\hline

\multicolumn{1}{l|}{Total} & \multicolumn{1}{l|}{16881} & 2690 & 76 & 55 & 898 & 3758 & 69 & 787 & 392 & 1547 & 451 & 202 & 184 & 283 & 66 & 152 & 5271 \\

\hline

\end{tabular}
\label{ShapeNet_Dataset}
\end{table*}

The GAP computes the mean of the transformed feature maps, $F^{\prime}$,  to obtain the global feature $G$. For segmenting the 3D object, it is necessary that both the local and global information is preserved. To do so, we concatenate the per point local feature map from $F^{\prime}$ with the global feature $G$ to obtain the concatenated per point features $C = \{(f^{\prime}_1, G), (f^{\prime}_2, G), ..., (f^{\prime}_n, G)\}$. Thus, the network becomes aware of both the local and global knowledge of the point clouds which intensifies the ability of the network to predict the appropriate per point labels.

Finally, the concatenated features $C$ is fed to the convolution layers and combine them to construct the output. For every point in the point cloud, the network predicts the probability of the point belonging to a particular part, thus generating the per point scores as the output.


\begin{table*}[ht]
\footnotesize
\setlength{\tabcolsep}{2pt}
\centering
\caption{Dataset statistics of PartNet dataset. The notations L1, L2 and L3 refer to the three levels of segmentation, namely, coarse level, middle level and fine-grained level.}
\begin{tabular}{l*{27}{l}l}
\hline

\multicolumn{1}{l|}{} & \multicolumn{1}{c|}{Total} & bag & bed & bott & bowl & chair & clock & dish & disp & door & ear & fauc & hat & key & knife & lamp & lap & micro & mug & frid & scis & stora & table & trash & vase\\

\hline

\multicolumn{1}{l|}{\# parts (L1)} & \multicolumn{1}{l|}{128} & 4 & 4 & 6 & 4 & 6 & 6 & 3 & 3 & 3 & 6 & 8 & 6 & 3 & 5 & 18 & 3 & 3 & 4 & 3 & 3 & 7 & 11 & 5 & 4 \\

\hline

\multicolumn{1}{l|}{\# parts (L2)} & \multicolumn{1}{l|}{149} & - & 10 & - & - & 30 & - & 5 & - & 4 & - & - & - & - & - & 28 & - & 5 & - & 6 & - & 19 & 42 & - & - \\

\hline

\multicolumn{1}{l|}{\# parts (L3)} & \multicolumn{1}{l|}{268} & - & 15 & 9 & - & 39 & 11 & 7 & 4 & 5 & 10 & 12 & - & - & 10 & 41 & - & 6 & - & 7 & - & 24 & 51 & 11 & 6 \\

\hline

\multicolumn{1}{l|}{\# train} & \multicolumn{1}{l|}{18112} & 92 & 133 & 315 & 131 & 4489 & 406 & 111 & 633 & 149 & 147 & 435 & 170 & 111 & 221 & 1554 & 306 & 133 & 138 & 136 & 45 & 1588 & 5707 & 221 & 741\\

\multicolumn{1}{l|}{\# validation} & \multicolumn{1}{l|}{2646} & 5 & 24 & 37 & 18 & 617 & 50 & 19 & 104 & 25 & 28 & 81 & 16 & 41 & 29 & 234 & 45 & 12 & 19 & 20 & 10 & 230 & 843 & 37 & 102\\

\multicolumn{1}{l|}{\# test} & \multicolumn{1}{l|}{5169} & 29 & 37 & 84 & 39 & 1217 & 98 & 51 & 191 & 51 & 53 & 132 & 45 & 31 & 77 & 419 & 82 & 39 & 35 & 31 & 13 & 451 & 1668 & 63 & 233\\

\hline

\multicolumn{1}{l|}{Total} & \multicolumn{1}{l|}{25927} & 126 & 194 & 436 & 188 & 6323 & 554 & 181 & 928 & 225 & 228 & 648 & 231 & 183 & 327 & 2207 & 433 & 184 & 192 & 187 & 68 & 2269 & 8218 & 321 & 1076\\

\hline

\end{tabular}
\label{PartNet_Dataset}
\end{table*}

\subsection{Implementation details} The proposed PIG-Net architecture is implemented on Intel\textsuperscript{\textregistered} Xeon(R) processor at $3.5$ GHz and $128$ GB RAM. For segmentation, we set the batch size as $64$ and use the cross entropy loss function. The Adam optimizer, which is an extension to stochastic gradient descent is used to minimize the cross entropy loss with a learning rate of $0.001$. We implemented the segmentation network using Keras  library~\cite{Keras:2015} and train each category separately using Nvidia Quadro M2000. 

\subsection{Performance evaluation} The evaluation metric used to analyze the performance of PIG-Net is mean intersection over union (mIoU). For each object of category, to compute the mIoU of the object, the IoU between the ground truth and prediction is calculated for each part in the category. The IoUs for all parts are averaged in category to get the mIoU for that object. To calculate mIoU for the category,  take the average of mIoUs for all the objects in that category. We report the instance mIoU (Ins. mIoU) that is the average of IoUs of all the point cloud instances and category mIoU (Cat. mIoU) that is the average of mIoUs of all the categories.

\section{Experiments and Results}
\label{results}

In this section, we present the experimental results of the proposed PIG-Net on two state-of-the-art datasets (Section~\ref{Dataset}). We describe the data augmentation techniques used in Section~\ref{augmentation}. The experimental results and comparative analysis is provided in Sections~\ref{Experiments_ShapeNet} and~\ref{Experiments_PartNet}. We evaluate the effectiveness of our network by performing ablation study in Section~\ref{ablation_study}.

\noindent

\begin{table*}[ht]
\footnotesize
\setlength{\tabcolsep}{3.8pt}
\centering
\caption{Segmentation results on ShapeNet-part dataset. Evaluation metric is mIoU(\%).}
\begin{tabular}{l*{17}{l}l}
\hline

\multicolumn{1}{l|}{} & \multicolumn{1}{c|}{Cat.} & \multicolumn{1}{c|}{Ins.} & aero & bag & cap & car & chair & ear & guitar & knife & lamp & laptop & motor & mug & pistol & rocket & skate & table\\
\multicolumn{1}{l|}{} & \multicolumn{1}{c|}{mIoU} & \multicolumn{1}{c|}{mIoU} &  &  &  &  &  & phone &  &  &  &  & bike &  &  &  & board & \\

\hline


\hline

\multicolumn{1}{l|}{Yi~\cite{Yi:2016:ShapeNet}} & \multicolumn{1}{c|}{79.0} & \multicolumn{1}{c|}{81.4} & 81.0 & 78.4 & 77.7 & 75.7 & 87.6 & 61.9 & 92.0 & 85.4 & 82.5 & 95.7 & 70.6 & 91.9 & 85.9 & 53.1 & 69.8 & 75.3 \\

\multicolumn{1}{l|}{PointNet~\cite{Qi:2017:PointNet}} & \multicolumn{1}{c|}{80.4} & \multicolumn{1}{c|}{83.7} & 83.4 & 78.7 & 82.5 & 74.9 & 89.6 & 73.0 & 91.5 & 85.9 & 80.8 & 95.3 & 65.2 & 93.0 & 81.2 & 57.9 & 72.8 & 80.6 \\

\multicolumn{1}{l|}{PointNet++~\cite{Qi:2017:PointNet++}} & \multicolumn{1}{c|}{81.9} & \multicolumn{1}{c|}{85.1} & 82.4 & 79.0 & 87.7 & 77.3 & 90.8 & 71.8 & 91.0 & 85.9 & 83.7 & 95.3 & 71.6 & 94.1 & 81.3 & 58.7 & 76.4 & 82.6 \\

\multicolumn{1}{l|}{Kd-Net~\cite{Klokov:2017:KdNet}} & \multicolumn{1}{c|}{77.4} & \multicolumn{1}{c|}{82.3} & 80.1 & 74.6 & 74.3 & 70.3 & 88.6 & 73.5 & 90.2 & 87.2 & 81.0 & 94.9 & 57.4 & 86.7 & 78.1 & 51.8 & 69.9 & 80.3 \\

\multicolumn{1}{l|}{SpecCNN~\cite{Yi:2017:SpecCNN}} & \multicolumn{1}{c|}{82.0} & \multicolumn{1}{c|}{84.7} & 81.6 & 81.7 & 81.9 & 75.2 & 90.2 & 74.9 & 93.0 & 86.1 & 84.7 & 95.6 & 66.7 & 92.7 & 81.6 & 60.6 & 82.9 & 82.1 \\

\multicolumn{1}{l|}{O-CNN~\cite{Wang:2017:OCNN}} & \multicolumn{1}{c|}{82.2} & \multicolumn{1}{c|}{85.9} & 85.5 & \textbf{87.1} & 84.7 & 77.0 & 91.1 & \textbf{85.1} & 91.9 & 87.4 & 83.3 & 95.4 & 56.9 & \textbf{96.2} & 81.6 & 53.5 & 74.1 & 84.4 \\

\multicolumn{1}{l|}{PointGrid~\cite{Le:2018:PointGrid}} & \multicolumn{1}{c|}{82.2} & \multicolumn{1}{c|}{86.4} & \textbf{85.7} & 82.5 & 81.8 & 77.9 & \textbf{92.1} & 82.4 & 92.7 & 85.8 & 84.2 & 95.3 & 65.2 & 93.4 & 81.7 & 56.9 & 73.5 & 84.6 \\

\multicolumn{1}{l|}{KCNet~\cite{Shen:2018:KCNet}} & \multicolumn{1}{c|}{82.2} & \multicolumn{1}{c|}{84.7} & 82.8 & 81.5 & 86.4 & 77.6 & 90.3 & 76.8 & 91.0 & 87.2 & 84.5 & 95.5 & 69.2 & 94.4 & 81.6 & 60.1 & 75.2 & 81.3\\

\multicolumn{1}{l|}{SO-Net~\cite{Li:2018:SONet} } & \multicolumn{1}{c|}{81.0} & \multicolumn{1}{c|}{84.9} & 82.8 & 77.8 & 88.0 & 77.3 & 90.6 & 73.5 & 90.7 & 83.9 & 82.8 & 94.8 & 69.1 & 94.2 & 80.9 & 53.1 & 72.9 & 83.0\\

\multicolumn{1}{l|}{SpiderCNN~\cite{Xu:2018:SpiderCNN}} & \multicolumn{1}{c|}{82.4} & \multicolumn{1}{c|}{85.3} & 83.5 & 81.0 & 87.2 & 77.5 & 90.7 & 76.8 & 91.1 & 87.3 & 83.3 & 95.8 & 70.2 & 93.5 & 82.7 & 59.7 & 75.8 & 82.8\\

\multicolumn{1}{l|}{PointCNN~\cite{Li:2018:PointCNN}} & \multicolumn{1}{c|}{84.6} & \multicolumn{1}{c|}{86.1} & 84.1 & 86.5 & 86.0 & 80.8 & 90.6 & 79.7 & 92.3 & 88.4 & 85.3 & 96.1 & 77.2 & 95.3 & 84.2 & 64.2 & 80.0 & 82.3\\

\multicolumn{1}{l|}{MRTNet~\cite{Gadelha:2018:MRTNet}} & \multicolumn{1}{c|}{79.3} & \multicolumn{1}{c|}{83.0} & 81.0 & 76.7 & 87.0 & 73.8 & 89.1 & 67.6 & 90.6 & 85.4 & 80.6 & 95.1 & 64.4 & 91.8 & 79.7 & 57.0 & 69.1 & 80.6\\

\multicolumn{1}{l|}{RS-Net~\cite{Huang:2018:RS-Net}} & \multicolumn{1}{c|}{81.4} & \multicolumn{1}{c|}{84.9} & 82.7 & 86.4 & 84.1 & 78.2 & 90.4 & 69.3 & 91.4 & 87.0 & 83.5 & 95.4 & 66.0 & 92.6 & 81.8 & 56.1 & 75.8 & 82.2\\

\multicolumn{1}{l|}{DGCNN~\cite{Wang:2019:DGCNN}} & \multicolumn{1}{c|}{82.3} & \multicolumn{1}{c|}{85.2} & 84.0 & 83.4 & 86.7 & 77.8 & 90.6 & 74.7 & 91.2 & 87.5 & 82.8 & 95.7 & 66.3 & 94.9 & 81.1 & 63.5 & 74.5 & 82.6\\

\multicolumn{1}{l|}{KPConv deform~\cite{Thomas:2019:KPConv}} & \multicolumn{1}{c|}{85.1} & \multicolumn{1}{c|}{86.4} & 84.6 & 86.3 & 87.2 & 81.1 & 91.1 & 77.8 & 92.6 & 88.4 & 82.7 & 96.2 & 78.1 & 95.8 & 85.4 & 69.0 & 82.0 & 83.6\\

\multicolumn{1}{l|}{3D-GCN~\cite{Lin:2020:3D-GCN}} & \multicolumn{1}{c|}{82.1} & \multicolumn{1}{c|}{85.1} & 83.1 & 84.0 & 86.6 & 77.5 & 90.3 & 74.1 & 90.9 & 86.4 & 83.8 & 95.6 & 66.8 & 94.8 & 81.3 & 59.6 & 75.7 & 82.8\\

\multicolumn{1}{l|}{SGPN~\cite{Wang:2018:SGPN}} & \multicolumn{1}{c|}{82.8} & \multicolumn{1}{c|}{85.8} & 80.4 & 78.6 & 78.8 & 71.5 & 88.6 & 78.0 & 90.9 & 83.0 & 78.8 & 95.8 & 77.8 & 93.8 & \textbf{87.4} & 60.1 & 92.3 & 89.4\\

\multicolumn{1}{l|}{DensePoint~\cite{Liu:2019:DensePoint}} & \multicolumn{1}{c|}{84.2} & \multicolumn{1}{c|}{86.4} & 84.0 & 85.4 & \textbf{90.0} & 79.2 & 91.1 & 81.6 & 91.5 & 87.5 & 84.7 & 95.9 & 74.3 & 94.6 & 82.9 & 64.6 & 76.8 & 83.7\\


\multicolumn{1}{l|}{PAN~\cite{Pan:2019:PAN}} & \multicolumn{1}{c|}{82.6} & \multicolumn{1}{c|}{85.7} & 82.9 & 81.3 & 86.1 & 78.6 & 91.0 & 77.9 & 90.9 & 87.3 & 84.7 & 95.8 & 72.9 & 95.0 & 80.8 & 59.6 & 74.1 & 83.5\\

\multicolumn{1}{l|}{SFCNN~\cite{Rao:2019:SFCNN}} & \multicolumn{1}{c|}{82.7} & \multicolumn{1}{c|}{85.4} & 83.0 & 83.4 & 87.0 & 80.2 & 90.1 & 75.9 & 91.1 & 86.2 & 84.2 & 96.7 & 69.5 & 94.8 & 82.5 & 59.9 & 75.1 & 82.9\\

\multicolumn{1}{l|}{SPH3D-GCN~\cite{Lei:2020:SPH3D-GCN}} & \multicolumn{1}{c|}{84.9} & \multicolumn{1}{c|}{86.8} & 84.4 & 86.2 & 89.2 & \textbf{81.2} & 91.5 & 77.4 & 92.5 & 88.2 & 85.7 & \textbf{96.7} & \textbf{78.6} & 95.6 & 84.7 & 63.9 & 78.5 & 84.0\\
\hline

\multicolumn{1}{l|}{PIG-Net (Ours)} & \multicolumn{1}{c|}{\textbf{85.9}} & \multicolumn{1}{c|}{\textbf{90.5}} & 84.2 & 83.1 & 88.9 & 78.6 & 91.7 & 78.2 & \textbf{94.4} & \textbf{89.5} & \textbf{94.2} & 96.3 & 66.2 & 91.6 & 85.1 & \textbf{64.8} & \textbf{93.5} & \textbf{94.2}\\

\multicolumn{1}{l|}{PIGNetwithMaxPool} & \multicolumn{1}{c|}{82.6} & \multicolumn{1}{c|}{85.9} & 82.2 & 77.3 & 88.2 & 77.4 & 87.6 & 77.3 & 92.4 & 83.9 & 81.2 & 92.6 & 66.4 & 90.8 & 84.8 & 60.7 & 90.2 & 89.1\\
\hline

\end{tabular}
\label{ShapeNet_Results}
\end{table*}

\subsection{Data augmentation}
\label{augmentation}

We uniformly sample the point clouds to obtain 1024 points. During training, we augment the point cloud on-the-fly by randomly rotating the object along the up-axis. We augment the data using random anisotropic scaling (range: $[0.66, 1.5]$) and random translation (range: $[-0.2, 0.2]$), as in~\cite{Roman:2017:Range}. Additionally, jittering the position of each point by a Gaussian noise with zero mean and 0.01 standard deviation is done.

\subsection{Datasets}
\label{Dataset}

For experimental analysis, we use ShapeNet-part dataset~\cite{Yi:2016:ShapeNet} and the recently released PartNet v0 dataset~\cite{Mo:2019:PartNet}. We are the first to present the network analysis on PartNet. The sample point clouds from PartNet dataset are shown in Figure~\ref{samples}.

The ShapeNet-part dataset contains $16,881$ 3D objects spread across 16 categories. There are $50$ parts in total with $2$ to $6$ parts per category. The dataset is split into  $12,137$ objects for training, $1870$ for validation and $2874$ objects for testing~\cite{Yi:2016:ShapeNet}. The number of objects in training, validation and testing sets in each of the categories is shown in Table~\ref{ShapeNet_Dataset}. This is a challenging 
segmentation dataset since both the object categories and the object parts within the categories are highly imbalanced.

The PartNet dataset shared by~\cite{Mo:2019:PartNet} consists of fine-grained part annotations for $25,927$ objects distributed over 24 categories. It is a large-scale dataset of 3D objects with fine-grained, instance-level, and hierarchical 3D part annotations. The fine-grained semantic segmentation consists of three levels of segmentation, namely, coarse level, middle level and fine grained level. The number of objects in training, validation and testing sets in each of the categories is shown in Table~\ref{PartNet_Dataset}. Segmenting the fine-grained objects is a challenging task since it requires distinguishing the small and similar semantic parts of the objects.

\begin{figure*}
\centering
\includegraphics[width=1\linewidth]{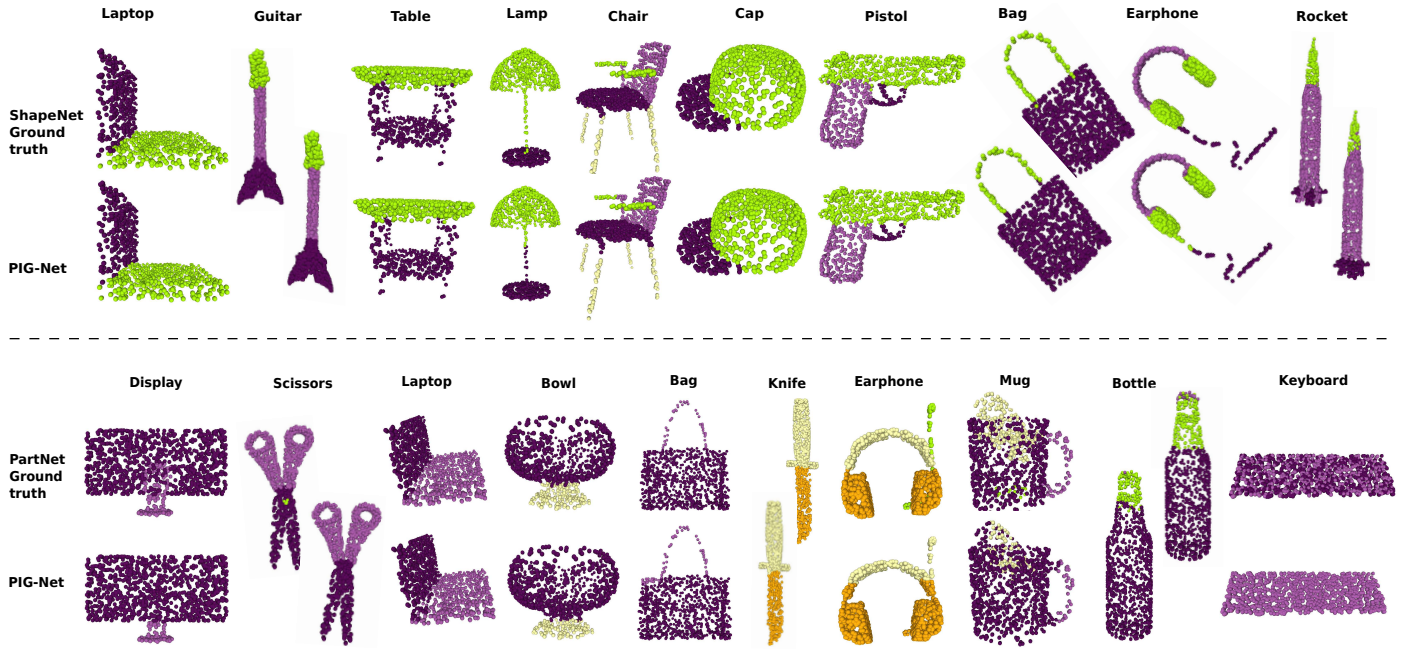}
\caption{Visualization of object part segmentation results using the proposed PIG-Net on ShapeNet-part dataset~\cite{Yi:2016:ShapeNet} (top row) and PartNet dataset~\cite{Mo:2019:PartNet} (bottom row).}
\label{Segmentation_Visualization}
\end{figure*}

\subsection{ShapeNet-part}
\label{Experiments_ShapeNet}

The results of the proposed PIG-Net and the existing state-of-the-art methods on ShapeNet-part dataset are shown in Table~\ref{ShapeNet_Results}. We can observe that the proposed approach achieves an improvement of 1\% on category mIoU and 3.7\% on instance mIoU, thus significantly outperforming the existing SOTA methods. Our Inception based PIG-Net architecture outperforms the most complex point cloud architectures like   PointCNN~\cite{Li:2018:PointCNN}, DGCNN~\cite{Wang:2019:DGCNN}, KPConv~\cite{Thomas:2019:KPConv} and SPH3D-GCN~\cite{Lei:2020:SPH3D-GCN}. The proposed technique achieves the best performance in 6/16 categories. In individual categories, our method ranks the best in \textit{guitar}, \textit{knife}, \textit{lamp},  \textit{rocket}, \textit{skateboard} and \textit{table} classes; the second best in \textit{chair} and  \textit{laptop} classes and the third best in \textit{cap} class. We can observe that our method performs better when there is more data, in categories such as \textit{table}, \textit{lamp}, \textit{chair} etc. The visualization of the segmented point clouds using PIG-Net is shown in Figure~\ref{Segmentation_Visualization} (top row), ordered from highest to lowest mIoU (left to right). As we can see from the visual results, our architecture segments most of the point clouds correctly. The \textit{rocket} category attains the lowest mIoU of 64.8\%. Most of the points in \textit{fin} and \textit{nose} parts are wrongly segmented to the body part of the \textit{rocket}. This is largely due to the very minute distinction and the continuity that is present in these parts leading to confusion.  
We also show performance of PIGNet with an ablation study by considering max pooling instead of GAP in the last row on Table~\ref{ShapeNet_Results}. We observe that results using max pooling shows lower IoUs in comparison to IoUs using GAP articulating that unlike max pooling which obtains only global features, GAP enforces the correspondences between global features and parts of 3D point cloud.


\begin{table*}[ht]
\footnotesize
\setlength{\tabcolsep}{1.8pt}
\centering
\caption{Segmentation results on PartNet dataset (fine-grained). Evaluation metric is mIoU(\%). The notations P, P$^{+}$, S, C and PIG refers to PointNet~\cite{Qi:2017:PointNet}, PointNet++~\cite{Qi:2017:PointNet++}, SpiderCNN~\cite{Xu:2018:SpiderCNN}, PointCNN~\cite{Li:2018:PointCNN} and our proposed PIG-Net respectively. The numbers 1, 2 and 3 refer to the three levels of segmentation, namely, coarse level, middle level and fine-grained level. The levels that are not defined are indicated by short lines.}

\begin{tabular}{c*{27}{c}c}
\hline

\multicolumn{1}{l|}{} & \multicolumn{1}{c|}{Cat.} & \multicolumn{1}{c|}{Ins.} & bag & bed & bott & bowl & chair & clock & dish & disp & door & ear & fauc & hat & key & knife & lamp & lap & micro & mug & frid & scis & stora & table & trash & vase\\
\multicolumn{1}{l|}{} & \multicolumn{1}{c|}{mIoU} & \multicolumn{1}{c|}{mIoU} &  &  &  &  &  &  &  &  &  &  &  &  &  &  &  &  &  & &  &  & \\

\hline

%

\multicolumn{1}{l|}{P1} & \multicolumn{1}{c|}{57.9} & \multicolumn{1}{c|}{55.1} & 42.5 & 32.0 & 33.8 & 58.0 & 64.6 & 33.2 & 76.0 & 86.8 & \textbf{64.4} & 53.2 & 58.6 & 55.9 & 65.6 & 62.2 & 29.7 & 96.5 & 49.4 & 80.0 & 49.6 & 86.4 & 51.9 & 50.5 & 55.2 & 54.7\\
\multicolumn{1}{l|}{P2} & \multicolumn{1}{c|}{37.3} & \multicolumn{1}{c|}{36.2} & - & 20.1 & - & - & 38.2 & - & 55.6 & - & 38.3 & - & - & - & - & - & 27.0 & - & 41.7 & - & 35.5 & - & 44.6 & 34.3 & - & - \\
\multicolumn{1}{l|}{P3} & \multicolumn{1}{c|}{35.6} & \multicolumn{1}{c|}{30.5} & - & 13.4 & 29.5 & - & 27.8 & 28.4 & 48.9 & 76.5 & 30.4 & 33.4 & 47.6 & - & - & 32.9 & 18.9 & - & 37.2 & - & 33.5 & - & 38.0 & 29.0 & 34.8 & 44.4\\

\hline

\multicolumn{1}{l|}{Avg} & \multicolumn{1}{c|}{51.2} & \multicolumn{1}{c|}{43.4} & 42.5 & 21.8 & 31.7 & 58.0 & 43.5 & 30.8 & 60.2 & 81.7 & 44.4 & 43.3 & 53.1 & 55.9 & 65.6 & 47.6 & 25.2 & 96.5 & 42.8	& 80.0 & 39.5 & 86.4 & 44.8 & 37.9 & 45.0 & 49.6\\

\hline

\multicolumn{1}{l|}{P$^{+}$1} & \multicolumn{1}{c|}{65.5} & \multicolumn{1}{c|}{65.3} & 59.7 & 51.8 & 53.2 & 67.3 & 68.0 & 48.0 & 80.6 & 89.7 & 59.3 & 68.5 & 64.7 & 62.4 & 62.2 & 64.9 & 39.0 & 96.6 & 55.7 & 83.9 & 51.8 & 87.4 & 58.0 & 69.5 & 64.3 & 64.4\\
\multicolumn{1}{l|}{P$^{+}$2} & \multicolumn{1}{c|}{44.5} & \multicolumn{1}{c|}{42.3} & - & 38.8 & - & - & 43.6 & - & 55.3 & - & 49.3 & - & - & - & - & - & 32.6 & - & 48.2 & - & 41.9 & - & 49.6 & 41.1 & - & - \\
\multicolumn{1}{l|}{P$^{+}$3} & \multicolumn{1}{c|}{42.5} & \multicolumn{1}{c|}{38.9} & - & 30.3 & 41.4 & - & 39.2 & 41.6 & 50.1 & 80.7 & 32.6 & 38.4 & 52.4 & - & - & 34.1 & 25.3 & - & 48.5 & - & 36.4 & - & 40.5 & 33.9 & 46.7 & 49.8\\
\hline

\multicolumn{1}{l|}{Avg} & \multicolumn{1}{c|}{58.1} & \multicolumn{1}{c|}{51.2} & 59.7 & 40.3 & 47.3 & 67.3 & 50.3 & 44.8 & 62.0 & 85.2 & 47.1 & 53.5 & 58.6 & 62.4 & 62.2 & 49.5 & 32.3 & 96.6 & 50.8 & 83.9 & 43.4 & 87.4 & 49.4 & 48.2 & 55.5 & 57.1\\

\hline

\multicolumn{1}{l|}{S1} & \multicolumn{1}{c|}{60.4} & \multicolumn{1}{c|}{57.3} & 57.2 & 55.5 & 54.5 & 70.6 & 67.4 & 33.3 & 70.4 & 90.6 & 52.6 & 46.2 & 59.8 & 63.9 & 64.9 & 37.6 & 30.2 & \textbf{97.0} & 49.2 & 83.6 & 50.4 & 75.6 & 61.9 & 50.0 & 62.9 & 63.8\\
\multicolumn{1}{l|}{S2} & \multicolumn{1}{c|}{41.7} & \multicolumn{1}{c|}{37.0} & - & 40.8 & - & - & 39.6 & - & 59.0 & - & 48.1 & - & - & - & - & - & 24.9 & - & 47.6 & - & 34.8 & - & 46.0 & 34.5 & - & -\\
\multicolumn{1}{l|}{S3} & \multicolumn{1}{c|}{37.0} & \multicolumn{1}{c|}{15.7} & - & 36.2 & 32.2 & - & 30.0 & 24.8 & 50.0 & 80.1 & 30.5 & 37.2 & 44.1 & - & - & 22.2 & 19.6 & - & 43.9 & - & 39.1 & - & 44.6 & 20.1 & 42.4 & 32.4\\

\hline

\multicolumn{1}{l|}{Avg} & \multicolumn{1}{c|}{53.6} & \multicolumn{1}{c|}{43.9} & 57.2 & 44.2 & 43.4 & 70.6 & 45.7 & 29.1 & 59.8 & 85.4 & 43.7 & 41.7 & 52.0 & 63.9 & 64.9 & 29.9 & 24.9 & 97.0 & 46.9 & 83.6 & 41.4 & 75.6 & 50.8 & 34.9 & 52.7 & 48.1 \\

\hline

\multicolumn{1}{l|}{C1} & \multicolumn{1}{c|}{64.3} & \multicolumn{1}{c|}{48.6} & 66.5 & 55.8 & 49.7 & 61.7 & 69.6 & 42.7 & \textbf{82.4} & \textbf{92.2} & 63.3 & 64.1 & 68.7 & \textbf{72.3} & \textbf{70.6} & 62.6 & 21.3 & \textbf{97.0} & 58.7 & \textbf{86.5} & 55.2 & \textbf{92.4} & 61.4 & 17.3 & \textbf{66.8} & 63.4 \\
\multicolumn{1}{l|}{C2} & \multicolumn{1}{c|}{46.5} & \multicolumn{1}{c|}{38.5} & - & 42.6 & - & - & 47.4 & - & \textbf{65.1} & - & 49.4 & - & - & - & - & - & 22.9 & - & \textbf{62.2} & - & 42.6 & - & 57.2 & 29.1 & - & -\\
\multicolumn{1}{l|}{C3} & \multicolumn{1}{c|}{46.4} & \multicolumn{1}{c|}{37.4} & - & 41.9 & 41.8 & - & 43.9 & 36.3 & \textbf{58.7} & \textbf{82.5} & \textbf{37.8} & 48.9 & 60.5 & - & - & 34.1 & 20.1 & - & \textbf{58.2} & - & 42.9 & - & 49.4 & 21.3 & \textbf{53.1} & 58.9\\

\hline

\multicolumn{1}{l|}{Avg} & \multicolumn{1}{c|}{59.8} & \multicolumn{1}{c|}{44.2} & 66.5 & 46.8 & 45.8 & 61.7 & 53.6 & 39.5 & \textbf{68.7} & \textbf{87.4} & 50.2 & 56.5 & 64.6 & \textbf{72.3} & \textbf{70.6} & 48.4 & 21.4 & \textbf{97.0} & 59.7 & \textbf{86.5} & 46.9 & \textbf{92.4} & 56.0 & 22.6 & \textbf{60.0} & 61.2 \\

\hline





\multicolumn{1}{l|}{PIG1} & \multicolumn{1}{c|}{\textbf{69.6}} & \multicolumn{1}{c|}{\textbf{72.6}} & \textbf{67.6} & \textbf{62.6} & \textbf{57.2} & \textbf{75.9} & \textbf{76.6} & \textbf{67.7} & 77.9 & 84.2 & 63.3 & \textbf{71.2} & \textbf{73.4} & 70.7 & 49.7 & \textbf{71.7} & \textbf{48.7} & 78.2 & \textbf{76.9} & 75.1 & \textbf{62.9} & 81.6 & \textbf{63.0} & \textbf{78.8} & 63.6 & \textbf{72.7}\\

\multicolumn{1}{l|}{PIG2} & \multicolumn{1}{c|}{\textbf{55.2}} & \multicolumn{1}{c|}{\textbf{55.1}} & - & \textbf{45.9} & - & - & \textbf{50.4} & - & 52.4 & - & \textbf{60.8} & - & - & - & - & - & \textbf{49.0} & - & 53.3 & - & \textbf{64.0} & - & \textbf{63.2} & \textbf{57.9} & - & -\\

\multicolumn{1}{l|}{PIG3} & \multicolumn{1}{c|}{\textbf{50.4}} & \multicolumn{1}{c|}{\textbf{44.9}} & - & \textbf{51.1} & \textbf{46.1} & - & \textbf{46.7} & \textbf{45.2} & 52.1 & 74.3 & 34.3 & \textbf{58.3} & \textbf{68.9} & - & - & \textbf{36.8} & \textbf{29.9} & - & 54.5 & - & \textbf{51.3} & - & \textbf{57.3} & \textbf{35.2} & 49.5 & \textbf{64.8}\\

\hline

\multicolumn{1}{l|}{Avg} & \multicolumn{1}{c|}{\textbf{62.9}} & \multicolumn{1}{c|}{\textbf{59.0}} & \textbf{67.6} & \textbf{53.2} & \textbf{51.7} & \textbf{75.9} & \textbf{57.9} & \textbf{56.5} & 60.8 & 79.3 & \textbf{52.8} & \textbf{64.8} & \textbf{71.2} & 70.7 & 49.7 & \textbf{54.3} & \textbf{42.5} & 78.2 & \textbf{61.6} & 75.1 & \textbf{59.4} & 81.6 & \textbf{61.2} & \textbf{57.3} & 56.5 & \textbf{68.8}\\

\hline

\end{tabular}
\label{PartNet_Results}
\end{table*}

\begin{table*}[ht]
\scriptsize
\setlength{\tabcolsep}{1.8pt}
\centering
\caption{Comparison of alternative PIG-Net architectures on PartNet dataset (coarse-level) w.r.t. variations in number of inception and pooling layers. The $Inc3L$ is a $3$ layer Inception$(64, 128, 256)$, $Inc4L$ is a $4$ layer Inception$(64, 128, 256, 512)$, and $Inc5L$ is a $5$ layer Inception$(64, 128, 256, 512, 1024)$.}

\begin{tabular}{c*{27}{c}c}
\hline

\multicolumn{1}{l|}{} & \multicolumn{1}{c|}{Cat.} & \multicolumn{1}{c|}{Ins.} & bag & bed & bott & bowl & chair & clock & dish & disp & door & ear & fauc & hat & key & knife & lamp & lap & micro & mug & frid & scis & stora & table & trash & vase\\
\multicolumn{1}{l|}{} & \multicolumn{1}{c|}{mIoU} & \multicolumn{1}{c|}{mIoU} &  &  &  &  &  &  &  &  &  &  &  &  &  &  &  &  &  & &  &  & \\


\hline

\multicolumn{1}{l|}{PIGNet-$Inc3L$-GAP} & \multicolumn{1}{c|}{66.7} & \multicolumn{1}{c|}{71.3} & 64.7 & 62.2 & 55.0 & 72.8 & 75.7 & 66.9 & 69.0 & 78.1 & 63.2 & 69.1 & 65.2 & 65.9 & 40.7 & 63.5 & 48.2 & 75.3 & 76.9 & 73.0 & 59.7 & 79.7 & 64.1 & 78.3 & 64.2 & 71.2\\

\multicolumn{1}{l|}{\textbf{PIGNet-\textit{Inc4L}-GAP}} & \multicolumn{1}{c|}{\textbf{69.6}} & \multicolumn{1}{c|}{\textbf{72.6}} & 67.6 & 62.6 & 57.2 & 75.9 & 76.6 & 67.7 & 77.9 & 84.2 & 63.3 & 71.2 & 73.4 & 70.7 & 49.7 & 71.7 & 48.7 & 78.2 & 76.9 & 75.1 & 62.9 & 81.6 & 63.0 & 78.8 & 63.6 & 72.7\\

\multicolumn{1}{l|}{PIGNet-$Inc5L$-GAP} & \multicolumn{1}{c|}{68.0} & \multicolumn{1}{c|}{71.5} & 64.8 & 64.5 & 48.8 & 70.9 & 76.5 & 66.3 & 76.9 & 83.3 & 61.0 & 66.8 & 74.5 & 68.4 & 47.0 & 75.4 & 48.3 & 74.9 & 78.2 & 68.1 & 61.8 & 79.8 & 62.9 & 76.9 & 65.4 & 70.8\\

\multicolumn{1}{l|}{PIGNet-No-$Inc4L$-} & \multicolumn{1}{c|}{61.9} & \multicolumn{1}{c|}{62.5} & 58.3 & 54.2 & 47.5 & 70.3 & 66.1 & 54.7 & 74.3 & 75.1 & 61.2 & 60.7 & 63.3 & 60.5 & 49.8 & 61.4 & 39.1 & 76.2 & 65.3 & 72.8 & 53.9 & 77.2 & 56.1 & 66.2 & 56.2 & 64.3\\
\multicolumn{1}{l|}{GAP} & \multicolumn{1}{c|}{} & \multicolumn{1}{c|}{}\\


\hline


\multicolumn{1}{l|}{PIGNet-$Inc4L$-max } & \multicolumn{1}{c|}{66.1} & \multicolumn{1}{c|}{69.1} & 63.2 & 60.1 & 54.6 & 73.8 & 72.6 & 62.9 & 75.0 & 80.7 & 61.3 & 68.4 & 69.9 & 69.1 & 45.9 & 66.3 & 45.2 & 75.5 & 74.6 & 71.3 & 57.8 & 73.1 & 61.2 & 76.4 & 59.7 & 68.9\\
\multicolumn{1}{l|}{pooling} & \multicolumn{1}{c|}{} & \multicolumn{1}{c|}{}\\



\hline

\end{tabular}
\label{inception_maxpooling}
\end{table*}

\subsection{PartNet}
\label{Experiments_PartNet}

The recently released PartNet is the large-scale dataset. The results of the proposed PIG-Net and the existing state-of-the-art methods on PartNet dataset are shown in Table~\ref{PartNet_Results}. We can observe that our method performs well on all the three levels of the PartNet dataset. In coarse level, PIG-Net outperforms other methods with an improvement of 4.1\% on category mIoU and 7.3\% on instance mIoU. We obtain the best results in 15/24 categories. The worst performance is achieved for \textit{keyboard} class with mIoU of 49.7\%. In \textit{keyboard} ground truth, the background panel is a different part than the foreground keys. Due to the intra-part closeness of the keys and the panel, our network segments these parts into a single part (see \textit{keyboard} in Figure~\ref{Segmentation_Visualization}). In middle level, we achieve the best performance in seven of nine categories. In individual categories, our method ranks the best in \textit{bed}, \textit{chair}, \textit{door}, \textit{lamp}, \textit{fridge}, \textit{storage furniture} and \textit{table} categories with an improvement of 8.7\% on category mIoU and 12.8\% on instance mIoU. 
The visualization of the segmented point clouds using our PIG-Net is shown in Figure~\ref{Segmentation_Visualization} (bottom row). 
In fine-grained level, we achieve the best results in 12/17 categories with an improvement of 4\% on category mIoU and 6\% on instance mIoU. Overall, we beat the existing methods on all three levels of segmentation with an average improvement of 3.1\% on category mIoU and 7.8\% on instance mIoU. PartNet being a challenging dataset, the experimental results clearly illustrates the effectiveness of PIG-Net, especially in fine-grained and middle levels, as our method is able to recognize the subtle details of the point clouds, much better than the existing methods. This is mainly due to the discriminative and rich features extracted using the inception layers as well as the global features extracted using the GAP layer in the PIG-Net architecture.





\subsection{Ablation Study}
\label{ablation_study}
In this section, we evaluate the effect of PIG-Net w.r.t. different architectures, robustness tests and network complexity on PartNet (coarse-level) dataset~\cite{Mo:2019:PartNet}. 


\subsubsection{Alternative Network Architecture}
\label{alternative_architecture}
We conduct experiments with alternative PIG-Net architectures and report their individual, overall category and instance mIoU for the 3D object segmentation on PartNet dataset. The performance of PIG-Net with different inception layers \textit{i.e.,} 3, 4, and 5 is shown in Table~\ref{inception_maxpooling}. The proposed PIG-Net in Figure~\ref{architecture} is a 4 layer inception with (64, 128, 256, 512) filters which out performs other inception architectures using GAP. When the number of inception layers further increases, we observe that the performance of the network drops as shown in Table~\ref{inception_maxpooling}. This is because as the number of layers increases, the network grows deep and the learning is  difficult due to vanishing gradient. In addition, it might lead to overfitting due to increase in the number of layers and filters in the inception~\cite{Szegedy:2015:Inception} bringing the necessity of changing the inception architecture. 
We also observe that performance of PIG-Net is poor without inception.


As mentioned in Section~\ref{methodology}, the global features can be computed using max pooling aggregation which is adopted by most of the existing methods. We compare the max pooling with the GAP by fixing the rest of our architecture with 4 layer inception. From Table~\ref{inception_maxpooling}, we  observe that the GAP layer achieves the best performance by a significant margin (3.5\% on category and instance mIoU) demonstrating that it is a better choice as it effectively captures the global point cloud characteristics.

%
%

\subsubsection{Robustness test}
\label{robustness_test}

\begin{figure}
\centering
\includegraphics[height=1.8in, width=3.2in]{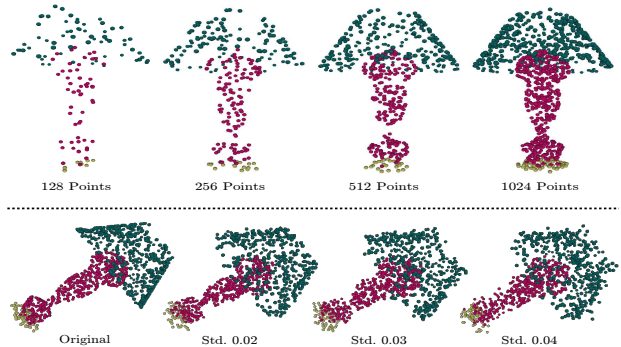}
\caption{Point cloud samples with different kinds of input corruptions. Top row: Sampling density variation. Bottom row: Addition of Gaussian noise.}
\label{Robustness_Samples}
\end{figure}

\begin{figure}
\centering
\includegraphics[height=1.1in, width=2.8in]{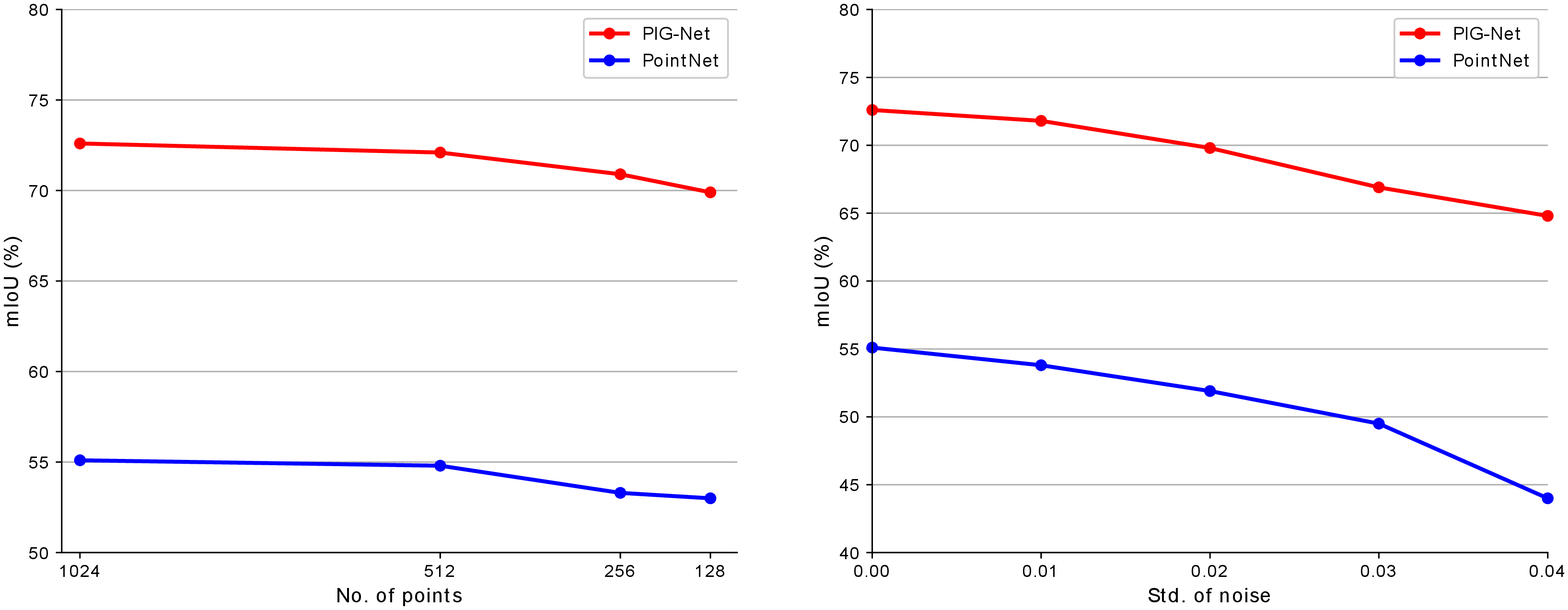}
\caption{Comparison of robustness test of the proposed PIG-Net and PointNet~\cite{Qi:2017:PointNet} architectures on PartNet dataset~\cite{Mo:2019:PartNet}. Left: Sampling density variation. Right: Addition of Gaussian noise.}
\label{Robustness_Plots}
\end{figure}

\textbf{Density variation:} We randomly sample the input points to 128, 256, 512, and 1024 as illustrated in Figure~\ref{Robustness_Samples} (top row). Figure~\ref{Robustness_Plots} (left) shows the instance mIoU's of the proposed PIG-Net and PointNet~\cite{Qi:2017:PointNet} architectures on PartNet dataset with varying number of input points. As to density variation, when the number of input points is decreased by 88\%, the performance drop of PIG-Net is only 2.7\%. The PIG-Net outperforms PointNet at all density levels.

\vspace{3pt}
\noindent
\textbf{Perturbation:} The Gaussian noise is added to each point independently as illustrated in Figure~\ref{Robustness_Samples} (bottom row). The standard deviation of the added noise is 0.01, 0.02, 0.03 and 0.04. Figure~\ref{Robustness_Plots} (right) shows instance mIoU's according to the different noise levels. The PIG-Net achieves about 65\% performance even when the point clouds are distorted by severe noise with the standard deviation of 0.04. The PIG-Net outperforms PointNet at all noise levels. These results indicate that the proposed PIG-Net is robust to various input corruptions and largely observed to be better than PointNet on complex PartNet dataset.

\subsubsection{Time and Space Complexity}
\label{time_space_complexity}

Table~\ref{time} summarizes space (number of parameters in the architecture) and time  (seconds) complexity w.r.t. learning model of PIG-Net architecture. In terms of space complexity, PIG-Net is more efficient than PointNet with $1.2$ times less number of parameters in the network. However, since inception dominates computation cost, the PIG-Net's time complexity is 30\% more expensive than PointNet.

\begin{table}[ht]
\small
\centering
\caption{Time and space complexity of the proposed PIG-Net and PointNet architectures for 3D object segmentation on PartNet (coarse level) dataset. The ``M" stands for million.}
\begin{tabular}{l*{6}{l}r}
\hline

& Space complexity & Time complexity \\
& (\#params) & (seconds)\\

\hline

PointNet~\cite{Qi:2017:PointNet} & 3.5M & 151.3s\\

PIG-Net (Ours) & 2.9M & 208.2s \\
\hline

\end{tabular}
\label{time}
\end{table}

\section{Conclusions and Future Directions}
\label{conclusions}

In this work, we proposed a inception based deep learning architecture, PIG-Net, suitable for 3D point cloud segmentation task. PIG-Net comprising of inception module and GAP, is invariant to geometric transformations, captures the local and global features, avoids overfitting and recognizes the fine-grained details of point clouds to enhance the performance. The experimental analysis of PIG-Net on widely popular challenging datasets, namely, ShapeNet~\cite{Yi:2016:ShapeNet} and PartNet~\cite{Mo:2019:PartNet} illustrates the effectiveness of our network. PIG-Net outperformed the existing state-of-the-art complex deep learning methods with an improvement of 1\% on category mIoU and 3.7\% on instance mIoU for ShapeNet-part and an improvement of 4.1\% on category mIoU and 7.3\% on instance mIoU for PartNet. With a boost in the performance of PIG-Net architecture by inception module and GAP, one of the promising future direction is to extend this to 3D scene understanding tasks.

\section*{Acknowledgement}

This research work is partly supported under the Indian Heritage in Digital Space (DST/ICPS/IHDS/2018) of Interdisciplinary Cyber Physical Systems  Programme of the Department of Science and Technology, Government of India.

\bibliographystyle{cag-num-names}
\bibliography{refs}

\end{document}